\documentclass[11pt,a4paper]{article}
\usepackage{emnlp2017}
\usepackage{times}
\usepackage{latexsym}

\usepackage{amsmath}
\usepackage{graphicx}
\usepackage{epstopdf}

\usepackage{booktabs} 

\usepackage{balance}

\usepackage{xspace}

\usepackage{bm}

\usepackage{array}

\usepackage{algorithm}
\usepackage[noend]{algpseudocode}

\usepackage{multirow}

\usepackage{enumitem}

\usepackage{subfigure} 	

\usepackage[font={small}]{caption}

\usepackage{breakurl}

\newenvironment{remark}[1][Remark]{\begin{trivlist}
\item[\hskip \labelsep {\bfseries #1}]}{\end{trivlist}}

\newcommand{\nop}[1]{}

\newenvironment{formalfont}{\fontfamily{pcr}\selectfont\small}{\par}
\DeclareTextFontCommand{\textformal}{\formalfont}

\newcommand{\Overnight}{\textsc{\small Overnight}\xspace}
\newcommand{\Calendar}{\textsc{\small Calendar}\xspace}
\newcommand{\Blocks}{\textsc{\small Blocks}\xspace}
\newcommand{\Housing}{\textsc{\small Housing}\xspace}
\newcommand{\Restaurants}{\textsc{\small Restaurants}\xspace}
\newcommand{\Publications}{\textsc{\small Publications}\xspace}
\newcommand{\Recipes}{\textsc{\small Recipes}\xspace}
\newcommand{\Social}{\textsc{\small Social}\xspace}
\newcommand{\Basketball}{\textsc{\small Basketball}\xspace}

\newcommand{\wordtovec}{\textsc{\small word2vec}\xspace}

\newcolumntype{L}[1]{>{\raggedright\let\newline\\\arraybackslash\hspace{0pt}}m{#1}}
\newcolumntype{C}[1]{>{\centering\let\newline\\\arraybackslash\hspace{0pt}}m{#1}}
\newcolumntype{R}[1]{>{\raggedleft\let\newline\\\arraybackslash\hspace{0pt}}m{#1}}

\DeclareMathOperator*{\argmax}{arg\,max}

\emnlpfinalcopy

\begin{document}

\title{Cross-domain Semantic Parsing via Paraphrasing}

\author{Yu Su \\
    Department of Computer Science \\
    University of California, Santa Barbara \\
  {\tt ysu@cs.ucsb.edu} \\
  \And
  Xifeng Yan \\
  Department of Computer Science \\
  University of California, Santa Barbara \\
  {\tt xyan@cs.ucsb.edu}
}

\maketitle

\begin{abstract}
Existing studies on semantic parsing mainly focus on the in-domain setting. We formulate cross-domain semantic parsing as a domain adaptation problem: train a semantic parser on some source domains and then adapt it to the target domain. Due to the diversity of logical forms in different domains, this problem presents unique and intriguing challenges. By converting logical forms into canonical utterances in natural language, we reduce semantic parsing to paraphrasing, and develop an attentive sequence-to-sequence paraphrase model that is general and flexible to adapt to different domains. We discover two problems, \emph{small micro variance} and \emph{large macro variance}, of pre-trained word embeddings that hinder their direct use in neural networks, and propose standardization techniques as a remedy. On the popular \Overnight dataset, which contains eight domains, we show that both cross-domain training and standardized pre-trained word embedding can bring significant improvement.
\end{abstract}

\section{Introduction}
\label{sec:introduction}

\begin{figure*}[t]
  \centering
  \includegraphics[width=.85\textwidth]{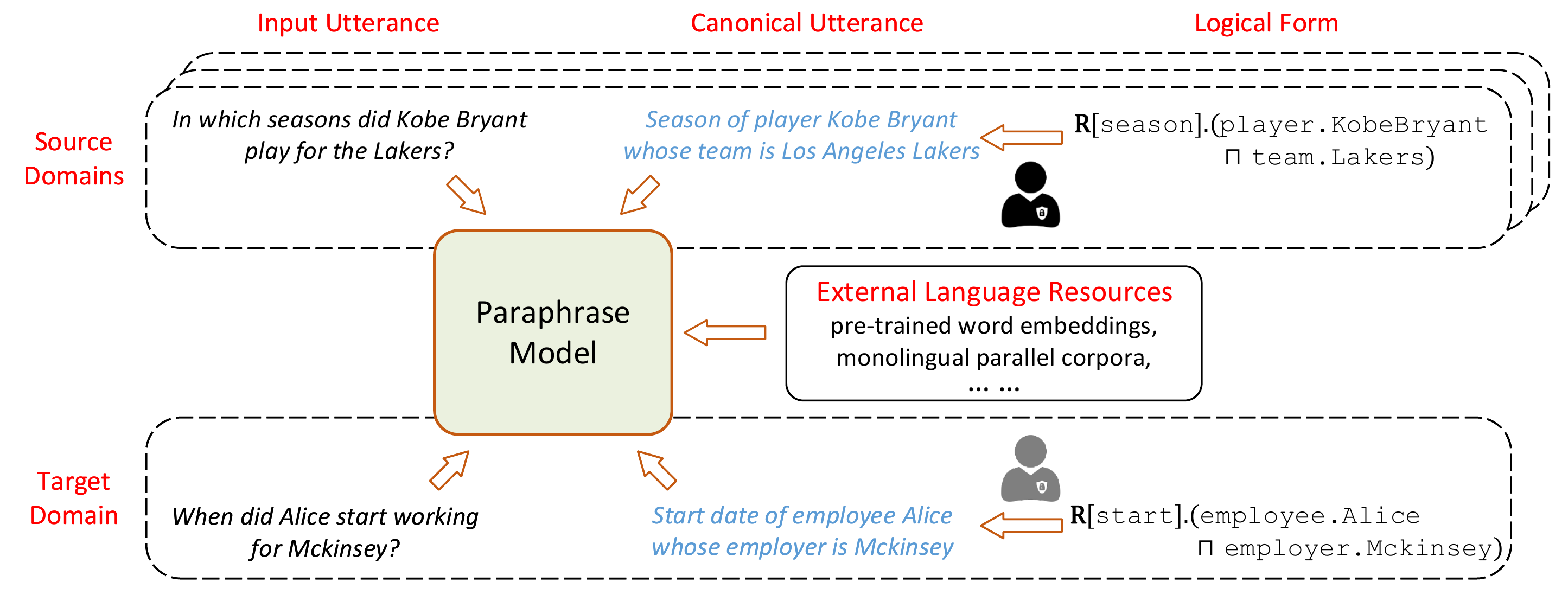}
  \caption{Cross-domain semantic parsing via paraphrasing framework. In a deterministic way, logical forms are first converted into canonical utterances in natural language. A paraphrase model then learns from the source domains and adapts to the target domain. External language resources can be incorporated in a consistent way across domains.}
  \label{fig:framework}
  \vspace{-5pt}
\end{figure*}

Semantic parsing, which maps natural language utterances into computer-understandable logical forms, has drawn substantial attention recently as a promising direction for developing natural language interfaces to computers.
Semantic parsing has been applied in many domains, including querying data/knowledge bases~\cite{woods1973progress, zelle1996learning, berant2013semantic}, controlling IoT devices~\cite{campagna2017almond}, and communicating with robots~\cite{chen2011learning, tellex2011understanding, artzi2013weakly, bisk2016natural}.

Despite the wide applications, studies on semantic parsing have mainly focused on the \emph{in-domain} setting, where both training and testing data are drawn from the same domain.
How to build semantic parsers that can learn across domains remains an under-addressed problem.
In this work, we study \emph{cross-domain semantic parsing}.
We model it as a domain adaptation problem~\cite{daume2006domain}, where we are given some \emph{source} domains and a \emph{target} domain, and the core task is to adapt a semantic parser trained on the source domains to the target domain (Figure~\ref{fig:framework}).
The benefits are two-fold:
(1) by training on the source domains, the cost of collecting training data for the target domain can be reduced, and
(2) the data of source domains may provide information complementary to the data collected for the target domain, leading to better performance on the target domain.

This is a very challenging task.
Traditional domain adaptation~\cite{daume2006domain,blitzer2006domain} only concerns natural languages, while semantic parsing concerns both natural and formal languages.
Different domains often involve different predicates.
In Figure~\ref{fig:framework}, from the source \Basketball domain a semantic parser can learn the semantic mapping from natural language to predicates like \textformal{team} and \textformal{season}, but in the target \Social domain it needs to handle predicates like \textformal{employer} instead.
Worse still, even for the same predicate, it is legitimate to use arbitrarily different predicate symbols, e.g., other symbols like \textformal{hired\_by} or even \textformal{predicate1} can also be used for the \textformal{employer} predicate, reminiscent of the symbol grounding problem~\cite{harnad1990symbol}.
Therefore, directly transferring the mapping from natural language to predicate symbols learned from source domains to the target domain may not be much beneficial.

Inspired by the recent success of paraphrasing based semantic parsing~\cite{berant2014semantic, wang2015building}, we propose to use natural language as an intermediate representation for cross-domain semantic parsing.
As shown in Figure~\ref{fig:framework}, logical forms are converted into canonical utterances in natural language, and semantic parsing is reduced to paraphrasing.
It is the knowledge of paraphrasing, at lexical, syntactic, and semantic levels, that will be transferred across domains.

Still, adapting a paraphrase model to a new domain is a challenging and under-addressed problem.
To give some idea of the difficulty, for each of the eight domains in the popular \Overnight~\cite{wang2015building} dataset, 30\% to 55\% of the words never occur in any of the other domains, a similar problem observed in domain adaptation for machine translation~\cite{daume11lexicaladapt}.
The paraphrase model therefore can get little knowledge for a substantial portion of the target domain from the source domains.
We introduce pre-trained word embeddings such as \wordtovec~\cite{mikolov2013distributed} to combat the vocabulary variety across domains.
Based on recent studies on neural network initialization, we conduct a statistical analysis of pre-trained word embeddings and discover two problems that may hinder their direct use in neural networks: \emph{small micro variance}, which hurts optimization, and \emph{large macro variance}, which hurts generalization.
We propose to \emph{standardize} pre-trained word embeddings, and show its advantages both analytically and experimentally.

On the \Overnight dataset, we show that cross-domain training under the proposed framework can significantly improve model performance.
We also show that, compared with directly using pre-trained word embeddings or normalization as in previous work, the proposed standardization technique can lead to about 10\% absolute improvement in accuracy.

\section{Cross-domain Semantic Parsing}
\label{sec:preliminaries}

\subsection{Problem Definition}
\label{sec:problem_definition}

Unless otherwise stated, we will use $u$ to denote input utterance, $c$ for canonical utterance, and $z$ for logical form.
We denote $\mathcal{U}$ as the set of all possible utterances.
For a domain, suppose $\mathcal{Z}$ is the set of logical forms, a semantic parser is a mapping $f\colon \mathcal{U} \rightarrow \mathcal{Z}$ that maps every input utterance to a logical form (a \textformal{null} logical form can be included in $\mathcal{Z}$ to reject out-of-domain utterances).

In cross-domain semantic parsing, we assume there are a set of $K$ source domains $\{  \mathcal{Z}_i \}_{i=1}^{K}$, each with a set of training examples $\{ (u^i_j, z^i_j) \}_{j=1}^{N_i}$.
It is in principle advantageous to model the source domains separately~\cite{daume2006domain}, which retains the possibility of separating domain-general information from domain-specific information, and only transferring the former to the target domain.
For simplicity, here we merge the source domains into a single domain $\mathcal{Z}_s$ with training data  $\{ (u_i, z_i) \}_{i=1}^{N_s}$.
The task is to learn a semantic parser $f\colon \mathcal{U} \rightarrow \mathcal{Z}_t$ for a target domain $\mathcal{Z}_t$, for which we have a set of training examples $\{ (u_i, z_i) \}_{i=1}^{N_t}$.
Some characteristics can be summarized as follows:

\begin{itemize}
  \item $\mathcal{Z}_t$ and $\mathcal{Z}_s$ can be totally disjoint.
  \item The input utterance distribution of the source and the target domains can be independent and differ remarkably.
  \item Typically $N_t \ll N_s$.
\end{itemize}

In the most general and challenging case, $\mathcal{Z}_t$ and $\mathcal{Z}_s$ can be defined using different formal languages.
Because of the lack of relevant datasets, here we restrain ourselves to the case where $\mathcal{Z}_t$ and $\mathcal{Z}_s$ are defined using the same formal language, e.g., $\lambda$-DCS~\cite{liang2013lambda} as in the \Overnight dataset.

\subsection{Framework}
\label{sec:framework}

Our framework follows the research line of semantic parsing via paraphrasing~\cite{berant2014semantic, wang2015building}.
While previous work focuses on the in-domain setting, we discuss its applicability and advantages in the cross-domain setting, and develop techniques to address the emerging challenges in the new setting.

\begin{remark}[Canonical utterance.]
We assume a \emph{one-to-one} mapping $g\colon \mathcal{Z} \rightarrow \mathcal{C}$, where $\mathcal{C} \subset \mathcal{U}$ is the set of canonical utterances.
In other words, every logical form will be converted into a unique canonical utterance deterministically (Figure~\ref{fig:framework}).
Previous work~\cite{wang2015building} has demonstrated how to design such a mapping, where a domain-general grammar and a domain-specific lexicon are constructed to automatically convert every logical form to a canonical utterance.
In this work, we assume the mapping is given\footnote{In the experiments we use the provided canonical utterances of the \Overnight dataset.}, and focus on the subsequent paraphrasing and domain adaptation problems.

This design choice is worth some discussion.
The grammar, or at least the lexicon for mapping predicates to natural language, needs to be provided by domain administrators.
This indeed brings an additional cost, but we believe it is reasonable and even necessary for three reasons:
(1) Only domain administrators know the predicate semantics the best, so it has to be them to reveal that by grounding the predicates to natural language (the symbol grounding problem~\cite{harnad1990symbol}).
(2) Otherwise, predicate semantics can only be learned from supervised training data of each domain, bringing a significant cost on data collection.
(3) Canonical utterances are understandable by average users, and thus can also be used for training data collection via crowdsourcing~\cite{wang2015building, su2016on}, which can amortize the cost.

Take comparatives as an example.  In logical forms, comparatives can be legitimately defined using arbitrarily different predicates in different domains, e.g., \textformal{<}, \textformal{smallerInSize}, or even predicates with an ambiguous surface form, like \textformal{lt}.
When converting logical form to canonical utterance, however, domain administrators have to choose common natural language expressions like ``\emph{less than}'' and ''\emph{smaller}'', providing a shared ground for cross-domain semantic parsing.
\end{remark}

\begin{remark}[Paraphrase model.]
In the previous work based on paraphrasing~\cite{berant2014semantic, wang2015building}, semantic parsers are implemented as log-linear models with hand-engineered domain-specific features (including paraphrase features).
Considering the recent success of representation learning for domain adaptation~\cite{glorot2011domain, chen2012marginalized}, we propose a paraphrase model based on the sequence-to-sequence (Seq2Seq) model~\cite{sutskever2014sequence}, which can be trained end to end without feature engineering.
We show that it outperforms the previous log-linear models by a large margin in the in-domain setting, and can easily adapt to new domains.
\end{remark}

\begin{remark}[Pre-trained word embeddings.]
An advantage of reducing semantic parsing to paraphrasing is that external language resources become easier to incorporate.
Observing the vocabulary variety across domains, we introduce pre-trained word embeddings to facilitate domain adaptation.
For the example in Figure~\ref{fig:framework}, the paraphrase model may have learned the mapping from ``play for'' to ``whose team is'' in a source domain.
By acquiring word similarities (``play''-``work'' and ``team''-``employer'') from pre-trained word embeddings, it can establish the mapping from ``work for'' to ``whose employer is'' in the target domain, even without in-domain training data.
We analyze statistical characteristics of the pre-trained word embeddings, and propose standardization techniques to remedy some undesired characteristics that may bring a negative effect to neural models.
\end{remark}

\begin{remark}[Domain adaptation protocol.]
We will use the following protocol: (1) train a paraphrase model using the data of the source domain, (2) use the learned parameters to initialize a model in the target domain, and (3) fine-tune the model using the training data of the target domain.
\end{remark}

\subsection{Prior Work}
\label{sec:related_work}

While most studies on semantic parsing so far have focused on the in-domain setting, there are a number of studies of particular relevance to this work.
In the recent efforts of scaling semantic parsing to large knowledge bases like Freebase~\cite{bollacker2008freebase}, researchers have explored several ways to infer the semantics of knowledge base relations unseen in training, which are often based on at least one (often both) of the following assumptions:
(1) \emph{Distant supervision}.
Freebase entities can be linked to external text corpora, and serve as anchors for seeking semantics of Freebase relations from text.
For example, Cai and Alexander~\shortcite{cai2013semantic}, among others~\cite{berant2013semantic, xu2016question}, use sentences from Wikipedia that contain any entity pair of a Freebase relation as the support set of the relation.
(2) \emph{Self-explaining predicate symbols}.
Most Freebase relations are described using a carefully chosen symbol (surface form), e.g., \textformal{place\_of\_birth}, which provides strong cues for their semantics.
For example, Yih et al.~\shortcite{yih2015semantic} directly compute the similarity of input utterance and the surface form of Freebase relations via a convolutional neural network.
Kwiatkowski et al.~\shortcite{kwiatkowski2013scaling} also extract lexical features from input utterance and the surface form of entities and relations.
They have actually evaluated their model on Freebase sub-domains not covered in training, and have shown impressive results.
However, in the more general setting of cross-domain semantic parsing, we may have neither of these luxuries.
Distant supervision may not be available (e.g., IoT devices involving no entities but actions), and predicate symbols may not provide enough cues (e.g., \textformal{predicate1}).
In this case, seeking additional inputs from domain administrators is probably necessary.

In parallel of this work, Herzig and Berant~\shortcite{herzig2017neural} have explored another direction of semantic parsing with multiple domains, where they use all the domains to train a single semantic parser, and attach a domain-specific encoding to the training data of each domain to help the semantic parser differentiate between domains.
We pursue a different direction: we train a semantic parser on some source domains and adapt it to the target domain.
Another difference is that their work directly maps utterances to logical forms, while ours is based on paraphrasing.

Cross-domain semantic parsing can be seen as a way to reduce the cost of training data collection, which resonates with the recent trend in semantic parsing.
Berant et al.~\shortcite{berant2013semantic} propose to learn from utterance-denotation pairs instead of utterance-logical form pairs, while Wang et al.~\shortcite{wang2015building} and Su et al.~\shortcite{su2016on} manage to employ crowd workers with no linguistic expertise for data collection.
Jia and Liang~\shortcite{jia2016data} propose an interesting form of data augmentation.
They learn a grammar from existing training data, and generate new examples from the grammar by recombining segments from different examples.

We use natural language as an intermediate representation to transfer knowledge across domains, and assume the mapping from the intermediate representation (canonical utterance) to logical form can be done deterministically.
Several other intermediate representations have also been used, such as combinatory categorial grammar~\cite{kwiatkowski2013scaling, reddy2014large}, dependency tree~\cite{reddy2016transforming,reddy2017universal}, and semantic role structure~\cite{goldwasser2013leveraging}.
But their main aim is to better represent input utterances with a richer structure.
A separate ontology matching step is needed to map the intermediate representation to logical form, which requires domain-dependent training.

A number of other related studies have also used paraphrasing.
For example, Fader et al.~\shortcite{fader2013paraphrase} leverage question paraphrases to for question answering, while Narayan et al.~\shortcite{narayan2016paraphrase} generate paraphrases as a way of data augmentation.

Cross-domain semantic parsing can greatly benefit from the rich literature of domain adaptation and transfer learning~\cite{daume2006domain, blitzer2006domain, pan2010survey, glorot2011domain}.
For example, Chelba and Acero~\shortcite{chelba2004adaptation} use parameters trained in the source domain as prior to regularize parameters in the target domain.
The feature augmentation technique from Daum{\'e} III~\shortcite{daume2009frustratingly} can be very helpful when there are multiple source domains.
We expect to see many of these ideas to be applied in the future.

\section{Paraphrase Model}
\label{sec:paraphrase_model}

In this section we propose a paraphrase model based on the Seq2Seq model~\cite{sutskever2014sequence} with soft attention.
Similar models have been used in semantic parsing~\cite{jia2016data,dong2016language} but for directly mapping utterances to logical forms.
We demonstrate that it can also be used as a paraphrase model for semantic parsing.
Several other neural models have been proposed for paraphrasing~\cite{socher2011dynamic, hu2014convolutional, yin2015multigrancnn}, but it is not the focus of this work to compare all the alternatives.

For an input utterance
$u=\left(u_1, u_2, \dots, u_m\right)$
and an output canonical utterance
$c=\left(c_1, c_2, \dots, c_n\right)$,
the model estimates the conditional probability
$p(c | u) = \prod_{j=1}^{n} p(c_j | u, c_{1:j-1} )$.
The tokens are first converted into vectors via a word embedding layer $\phi$.
The initialization of the word embedding layer is critical for domain adaptation, which we will further discuss in Section~\ref{sec:embedding_initialization}.

The \emph{encoder}, which is implemented as a bi-directional recurrent neural network (RNN), first encodes $u$ into a sequence of state vectors $\left(h_1, h_2, \dots, h_m\right)$.
The state vectors of the forward RNN and the backward RNN are respectively computed as:

\begin{equation*}\label{eq:encoder}
    \centering
    \begin{aligned}
        \overrightarrow{h}_i    &= GRU_{fw}(\phi(u_i), \overrightarrow{h}_{i-1}) \\
        \overleftarrow{h}_i      &= GRU_{bw}(\phi(u_i), \overleftarrow{h}_{i+1})
    \end{aligned}
\end{equation*}

\noindent where gated recurrent unit (GRU) as defined in \cite{cho2014learning} is used as the recurrence.
We then concatenate the forward and backward state vectors, $h_i = [\overrightarrow{h}_i, \overleftarrow{h}_i], i = 1, \dots, m$.

We use an attentive RNN as the \emph{decoder}, which will generate the output tokens one at a time.
We denote the state vectors of the decoder RNN as $\left(d_1, d_2, \dots, d_n\right)$.
The attention takes a form similar to \cite{vinyals2015grammar}.
For the decoding step $j$, the decoder is defined as follows:

\vspace{-15pt}
\begin{equation*}\label{eq:decoder}
    \centering
    \begin{aligned}
        d_0    &=    \tanh(W_0 [\overrightarrow{h}_m, \overleftarrow{h}_1]) \\
        u_{ji}  &=      v^T \tanh(W_1 h_i + W_2 d_j) \\
        \alpha_{ji} &= \frac{u_{ji}}{\sum_{i'=1}^{m} u_{ji'}} \\
        h'_{j} &= \sum_{i=1}^{m} \alpha_{ji} h_i \\
        d_{j+1} &= GRU( [\phi(c_j), h'_{j}], d_j ) \\
        p( c_j \vert u, c_{1:j-1} ) &\propto \exp( U [d_j, h'_j])
    \end{aligned}
\end{equation*}
\vspace{-10pt}

\noindent where $W_0, W_1, W_2, v$ and $U$ are model parameters.
The decoder first calculates normalized attention weights $\alpha_{ji}$ over encoder states, and get a summary state $h'_j$.
The summary state is then used to calculate the next decoder state $d_{j+1}$ and the output probability distribution $p( c_j \vert u, c_{1:j-1} )$.

\begin{remark}[Training.]
Given a set of training examples $\{ (u_i, c_i) \}_{i=1}^{N}$, we minimize the cross-entropy loss
$
  - \frac{1}{N} \sum_{i=1}^{N} \log p(c_i | u_i),
$
which maximizes the log probability of the correct canonical utterances.
We apply dropout~\cite{hinton2012improving} on both input and output of the GRU cells to prevent overfitting.
\end{remark}

\begin{remark}[Testing.]
Given a domain $\{  \mathcal{Z}, \mathcal{C} \}$, there are two ways to use a trained model.
One is to use it to \emph{generate} the most likely output utterance $u'$ given an input utterance $u$~\cite{sutskever2014sequence},

\vspace{-8pt}
\begin{equation*}\label{eq:test_generator}
  u' = \argmax_{u' \, \in \, \mathcal{U}} p(u' | u).
\end{equation*}
\vspace{-8pt}

\noindent In this case $u'$ can be any utterance permissable by the output vocabulary, and may not necessarily be a legitimate canonical utterance in $\mathcal{C}$.
This is more suitable for large domains with a lot of logical forms, like Freebase.
An alternative way is to use the model to \emph{rank} the legitimate canonical utterances~\cite{kannan2016smart}:

\vspace{-8pt}
\begin{equation*}\label{eq:test_generator}
  c = \argmax_{c \, \in \, \mathcal{C}} p(c | u),
\end{equation*}
\vspace{-8pt}

\noindent which is more suitable for small domains having a limited number of logical forms, like the ones in the \Overnight dataset.
We will adopt the second strategy.
It is also very challenging; random guessing leads to almost no success.
It is also possible to first find a smaller set of candidates to rank via beam search~\cite{berant2013semantic, wang2015building}.
\end{remark}

\section{Pre-trained Word Embedding for Domain Adaptation}
\label{sec:embedding_initialization}

Pre-trained word embeddings like \wordtovec have a great potential to combat the vocabulary variety across domains.
For example, we can use pre-trained \wordtovec vectors to initialize the word embedding layer of the source domain, with the hope that
the other parameters in the model will co-adapt with the word vectors during training in the source domain, and generalize better to the out-of-vocabulary words (but covered by \wordtovec) in the target domain.
However, deep neural networks are very sensitive to initialization~\cite{erhan2010does}, and a statistical analysis of the pre-trained \wordtovec vectors reveals some characteristics that may not be desired for initializing deep neural networks.
In this section we present the analysis and propose a standardization technique to remedy the undesired characteristics.

\begin{table}[t]
  \centering
  \small
  \scalebox{.85}{%
  \begin{tabular}{llll}
   \toprule
   Initialization       &       L2 norm         &   Micro Variance    &   Cosine Sim. \\
    \midrule
    Random              &   $17.3 \pm 0.45$    &  $1.00 \pm 0.05$       &  $0.00 \pm 0.06$         \\
    \wordtovec          &   $2.04 \pm 1.08$    &  $0.02 \pm 0.02$       &  $0.13 \pm 0.11$         \\
    \wordtovec + ES    &   $17.3 \pm 0.05$    &  $1.00 \pm 0.00$         &  $0.13 \pm 0.11$        \\
    \wordtovec + FS   &   $16.0 \pm 8.47$    &  $1.09 \pm 1.31$          &  $0.12 \pm 0.10$       \\
    \wordtovec + EN   &   $1.00 \pm 0.00$    &  $0.01 \pm 0.00$          &  $0.13 \pm 0.11$       \\
    \bottomrule
  \end{tabular}}
  \caption{Statistics of the word embedding matrix with different initialization strategies. Random: random sampling from $U(-\sqrt{3}, \sqrt{3})$, thus unit variance. \wordtovec: raw \wordtovec vectors. ES: per-example standardization. FS: per-feature standardization. EN: per-example normalization. Cosine similarity is computed on a randomly selected (but fixed) set of 1M word pairs.}
  \label{table:embedding_statistics}
  \vspace{-10pt}
\end{table}

\begin{remark}[Analysis.]
Our analysis will be based on the 300-dimensional \wordtovec vectors trained on the 100B-word Google News corpus\footnote{\burl{https://code.google.com/archive/p/word2vec/}}.
It contains 3 million words, leading to a 3M-by-300 word embedding matrix.
The ``rule of thumb'' to randomly initialize word embedding in neural networks is to sample from a uniform or Gaussian distribution with \emph{unit variance}, which works well for a wide range of neural network models in general.
We therefore use it as a reference to compare different word embedding initialization strategies.
Given a word embedding matrix, we compute the L2 norm of each row and report the mean and the standard deviation.
Similarly, we also report the variance of each row (denoted as \emph{micro variance}), which indicates how far the numbers in the row spread out, and pair-wise cosine similarity, which indicates the word similarity captured by \wordtovec.

The statistics of the word embedding matrix with different initialization strategies are shown in Table~\ref{table:embedding_statistics}.
Compared with random initialization, two characteristics of the \wordtovec vectors stand out: (1) \emph{Small micro variance}. Both the L2 norm and the micro variance of the \wordtovec vectors are much smaller. (2) \emph{Large macro variance}. The variance of different \wordtovec vectors, reflected by the standard deviation of L2 norm, is much larger (e.g., the maximum and the minimum L2 norm are 21.1 and 0.015, respectively).
Small micro variance can make the variance of neuron activations starts off too small\footnote{Under some conditions, including using Xavier initialization (also introduced in that paper and now widely used) for weights, Glorot and Bengio~\shortcite{glorot2010understanding} have shown that the activation variances in a feedforward neural network will be roughly the same as the input variances (word embedding here) at the beginning of training.}, implying a poor starting point in the parameter space.
On the other hand, because of the magnitude difference, large macro variance may make a model hard to generalize to words unseen in training.
\end{remark}

\begin{remark}[Standardization.]
Based on the above analysis, we propose to do \emph{unit variance standardization} (standardization for short) on pre-trained word embeddings.
There are two possible ways, \emph{per-example standardization}, which standardizes each row of the embedding matrix to unit variance by simply dividing by the standard deviation of the row, and \emph{per-feature standardization}, which standardizes each column instead.
We do not make the rows or columns zero mean.
Per-example standardization enjoys the goodness of both random initialization and pre-trained word embeddings: it fixes the small micro variance problem as well as the large macro variance problem of pre-trained word embeddings, while still preserving cosine similarity, i.e., word similarity.
Per-feature standardization does not preserve cosine similarity, nor does it fix the large macro variance problem.
However, it enjoys the benefit of \emph{global} statistics, in contrast to the \emph{local} statistics of individual word vectors used in per-example standardization.
Therefore, in problems where the testing and training vocabularies are similar, per-feature standardization may be more advantageous.
Both standardizations lose vector magnitude information.
Levy et al.~\shortcite{levy2015improving} have suggested \emph{per-example normalization}\footnote{It can also be found in the implementation of Glove~\cite{pennington2014glove}: \burl{https://github.com/stanfordnlp/GloVe}} of pre-trained word embeddings for lexical tasks like word similarity and analogy, which do no involve deep neural networks.
Making the word vectors unit length alleviates the large macro variance problem, but the small micro variance problem remains (Table~\ref{table:embedding_statistics}).
\end{remark}

\begin{remark}[Discussion.]
This is indeed a pretty simple trick, and per-feature standardization (with zero mean) is also a standard data preprocessing method.
However, it is not self-evident that this kind of standardization shall be applied on pre-trained word embeddings before using them in deep neural networks, especially with the obvious downside of rendering the word embedding algorithm's loss function sub-optimal.

We expect this to be less of a issue for large-scale problems with a large vocabulary and abundant training examples.
For example, Vinyals et al.~\shortcite{vinyals2015grammar} have found that directly using the \wordtovec vectors for initialization can bring a consistent, though small, improvement in neural constituency parsing.
However, for smaller-scale problems (e.g., an application domain of semantic parsing can have a vocabulary size of only a few hundreds), this issue becomes more critical.
Initialized with the raw pre-trained vectors, a model may quickly fall into a poor local optimum and may not have enough signal to escape.
Because of the large macro variance problem, standardization can be critical for domain adaptation, which needs to generalize to many words unseen in training.

The proposed standardization technique appears in a similar spirit to batch normalization~\cite{ioffe2015batch}.
We notice two computational differences, that ours is applied on the inputs while batch normalization is applied on internal neuron activations, and that ours standardizes the whole word embedding matrix beforehand while batch normalization standardizes each mini-batch on the fly.
In terms of motivation, the proposed technique aims to remedy some undesired characteristics of pre-trained word embeddings, and batch normalization aims to reduce the internal covariate shift.
It is of interest to study the combination of the two in future work.
\end{remark}

\begin{table*}[!ht]
  \centering
  \small
  \scalebox{0.8}{%
  \begin{tabular}{lllllllll}
    \toprule
    Metric                                  &   \Calendar   &   \Blocks   &   \Housing   &   \Restaurants   &   \Publications   &   \Recipes   &   \Social   &   \Basketball \\
    \midrule
    \# of example ($N$)                         &   837		       &    1995	    &   941		       &   1657		        &   801		         &   1080		    &  4419		&   1952 \\
    \# of logical form  ($\left| \mathcal{Z} \right|, \left| \mathcal{C} \right|$)                    &   196		       &    469	    &   231		       &   339		            &   149		         &   124		    &  624		    &   252 \\
    vocab. size  ($\left| \mathcal{V} \right|$)                      &   228		       &    227	    &   318		       &   342		            &   203		         &   256		    &  533		    &   360 \\
    \% $\in$ other domains &  71.1 &    61.7	 &   60.7		&   55.8	 &   65.6		     &   71.9		&  46.0    &   45.6 \\
    \% $\in$ \wordtovec  &   91.2		&    91.6	 &   88.4		&   88.6		     &   91.1		     &   93.8		&  86.9    &   86.9 \\
    \% $\in$ other domains + \wordtovec  &  \textbf{93.9} &   \textbf{93.8}	& \textbf{90.9} &  \textbf{90.4} &   \textbf{95.6}		     &  \textbf{97.3}		&  \textbf{89.3}    &   \textbf{89.4} \\
    \bottomrule
  \end{tabular}}
  \caption{Statistics of the domains in the \Overnight dataset. Pre-trained \wordtovec embedding covers most of the words in each domain, paving a way for domain adaptation.}
  \label{table:exp_data_statistics}
  \vspace{-5pt}
\end{table*}

\begin{table*}[!ht]
  \centering
  \small
  \scalebox{0.8}{%
  \begin{tabular}{llllllllll}
    \toprule
    Method  &   \Calendar   &   \Blocks   &   \Housing   &   \Restaurants   &   \Publications   &   \Recipes   &   \Social   &   \Basketball &   Avg.   \\
    \midrule
    \textbf{Previous Methods}   & &  &  &  &  & & & & \\
    Wang et al.~\shortcite{wang2015building}       &   74.4		      &    41.9	   &   54.0		     &   75.9		          &   59.0		           &   70.8		  &  48.2		  &   46.3          &   58.8 \\
    Xiao et al.~\shortcite{xiao2016sequence}     &   75.0	          &    55.6   &   61.9		      &  80.1		          &   75.8		           &   --	  &  80.0		  &   80.5              & 72.7 \\
    Jia and Liang~\shortcite{jia2016data}   &   78.0	     & 58.1     &    71.4	   &  76.2    &    76.4		      &   79.6		          & 81.4	           &   85.2		  &  75.8	\\
    Herzig and Berant~\shortcite{herzig2017neural}   & 82.1  &   \textbf{62.7}    &   78.3    &   82.2    &   \textbf{80.7}    &   82.9    &   81.7    &   86.2    &   79.6 \\

  \midrule
  \textbf{Our Methods}   & &  &  &  &  & & & & \\
  Random + I &   75.6	& 60.2	&  67.2	&  77.7	&  77.6	&  80.1	&  80.7	&  86.5	&  75.7  \\
  Random + X &  79.2	& 54.9	&  74.1	&  76.2	&  78.5	&  82.4	&  82.5	&  86.7	&  76.9  \\
  \midrule
  \wordtovec + I  &  67.9    &	 59.4	&  52.4	&  75.0	&  64.0	&  73.2	&  77.0	&  87.5	&  69.5\\
  \wordtovec + X &  78.0	&  54.4	&  63.0	&  81.3	&  74.5	&  83.3	&  81.5	&  83.1	&  74.9\\
  \midrule
   \wordtovec + EN + I &  63.1	& 56.1	& 60.3	&  75.3	&  65.2	&  69.0	&  76.4	&  81.8	&  68.4\\
   \wordtovec + EN + X & 78.0	& 52.6	& 63.5	&  74.7	&  65.2	&  80.6	&  79.9	&  80.8	&  71.2\\
  \midrule
   \wordtovec + FS + I  & 78.6	& 62.2	&  67.7	&  78.6	&  75.8	&  85.7	&  81.3	&  86.7	&  77.1\\
   \wordtovec + FS + X & \textbf{82.7}	& 59.4	&  75.1	&  80.4	&  78.9	&  85.2	&  81.8	&  87.2	&  78.9\\
  \midrule
   \wordtovec + ES + I &   79.8     & 60.2	   &  71.4	&  81.6   &   78.9    &   84.7		  &  82.9   &  86.2   &  78.2  \\
   \wordtovec + ES + X &  82.1     & 62.2	   &  \textbf{78.8}	&  \textbf{83.7}	 &   80.1    &  \textbf{86.1}		  &  \textbf{83.1}	  &  \textbf{88.2}   &   \textbf{80.6} \\
  \bottomrule
  \end{tabular}
  } %
  \caption{Main experiment results. We combine the proposed paraphrase model with different word embedding initializations. I: in-domain, X: cross-domain, EN: per-example normalization, FS: per-feature standardization, ES: per-example standardization.}
  \label{table:exp_overnight_overall}
  \vspace{-10pt}
\end{table*}

\section{Evaluation}
\label{sec:evaluation}

\subsection{Data Analysis}
\label{sec:exp_dataset}

The \Overnight dataset~\cite{wang2015building} contains 8 different domains.
Each domain is based on a separate knowledge base, with logical forms written in $\lambda$-DCS~\cite{liang2013lambda}.
Logical forms are converted into canonical utterances via a simple grammar, and the input utterances are collected by asking crowd workers to paraphrase the canonical utterances.
Different domains are designed to stress different types of linguistic phenomena.
For example, the \Calendar domain requires a semantic parser to handle temporal language like ``\emph{meetings that start after 10 am}'', while the \Blocks domain features spatial language like ``\emph{which block is above block 1}''.

Vocabularies vary remarkably across domains (Table~\ref{table:exp_data_statistics}).
For each domain, only 45\% to 70\% of the words are covered by any of the other 7 domains.
A model has to learn the out-of-vocabulary words from scratch using in-domain training data.
The pre-trained \wordtovec embedding covers most of the words of each domain, and thus can connect the domains to facilitate domain adaptation.
Words that are still missing are mainly stop words and typos, e.g., ``\emph{ealiest}''.

\subsection{Experiment Setup}
\label{sec:exp_setup}

We compare our model with all the previous methods evaluated on the \Overnight dataset.
Wang et al.~\shortcite{wang2015building} use a log-linear model with a rich set of features, including paraphrase features derived from PPDB~\cite{ganitkevitch2013ppdb}, to rank logical forms.
Xiao et al.~\shortcite{xiao2016sequence} use a multi-layer perceptron to encode the unigrams and bigrams of the input utterance, and then use a RNN to predict the derivation sequence of a logical form under a grammar.
Similar to ours, Jia and Liang~\shortcite{jia2016data} also use a Seq2Seq model with bi-directional RNN encoder and attentive decoder, but it is used to predict linearized logical forms.
They also propose a data augmentation technique, which further improves the average accuracy to 77.5\%.
But it is orthogonal to this work and can be incorporated in any model including ours, therefore not included.

The above methods are all based on the in-domain setting, where a separate parser is trained for each domain.
In parallel of this work, Herzig and Berant~\shortcite{herzig2017neural} have explored another direction of cross-domain training: they use all of the domains to train a single parser, with a special domain encoding to help differentiate between domains.
We instead model it as a domain adaptation problem, where training on the source and the target domains are separate.
Their model is the same as Jia and Liang~\shortcite{jia2016data}.
It is the current best-performing method on the \Overnight dataset.

We use the standard 80\%/20\% split of training and testing, and randomly hold out 20\% of training for validation.
In cross-domain experiments, for each target domain, all the other domains are combined as the source domain.
Hyper-parameters are selected based on the validation set.
State size of both the encoder and the decoder are set to 100, and word embedding size is set to 300.
Input and output dropout rate of the GRU cells are 0.7 and 0.5, respectively, and mini-batch size is 512.
We use Adam with the default parameters suggested in the paper for optimization.
We use gradient clipping with a cap for global norm at 5.0 to alleviate the exploding gradients problem of recurrent neural networks.
Early stopping based on the validation set is used to decide when to stop training.
The selected model is retrained using the whole training set (training + validation).
The evaluation metric is accuracy, i.e., the proportion of testing examples for which the top prediction yields the correct denotation.
Our model is implemented in Tensorflow~\cite{abadi2016tensorflow}, and the code can be found at \burl{https://github.com/ysu1989/CrossSemparse}.

\subsection{Experiment Results}
\label{sec:exp_results}

\subsubsection{Comparison with Previous Methods}
\label{sec:exp_main_exp}

The main experiment results are shown in Table~\ref{table:exp_overnight_overall}.
Our base model (Random + I) achieves an accuracy comparable to the previous best in-domain model~\cite{jia2016data}.
With our main novelties, cross-domain training and word embedding standardization, our full model is able to outperform the previous best model, and achieve the best accuracy on 6 out of the 8 domains.
Next we examine the novelties separately.

\subsubsection{Word Embedding Initialization}
The in-domain results clearly show the sensitivity of model performance to word embedding initialization.
Directly using the raw \wordtovec vectors or with per-example normalization, the performance is significantly worse than random initialization (6.2\% and 7.3\%, respectively).
Based on the previous analysis, however, one should not be too surprised.
The small micro variance problem hurts optimization.
In sharp contrast, both of the proposed standardization techniques lead to better in-domain performance than random initialization (1.4\% and 2.5\%, respectively), setting a new best in-domain accuracy (78.2\%) on \Overnight.
The results show that the pre-trained \wordtovec vectors can indeed provide useful information, but only when they are properly standardized.

\subsubsection{Cross-domain Training}
A consistent improvement from cross-domain training is observed across all word embedding initialization strategies.
Even for raw \wordtovec embedding or per-example normalization, cross-domain training helps the model escape the poor initialization, though still inferior to the alternative initializations.
The best results are again obtained with standardization, with per-example standardization bringing a slightly larger improvement than per-feature standardization.
We observe that the improvement from cross-domain training is correlated with the abundance of the in-domain training data of the target domain.
To further examine this observation, we use the ratio between the number of examples ($N$) and the vocabulary size ($\left| \mathcal{V} \right|$) to indicate the data abundance of a domain (the higher, the more abundant), and compute the Pearson correlation coefficient between data abundance and accuracy improvement from cross-domain training (X$-$I).
The results in Table~\ref{table:exp_correlation} show a consistent, moderate to strong negative correlation between the two variables.
In other words, cross-domain training is more beneficial when in-domain training data is less abundant, which is reasonable because in that case the model can learn more from the source domain data that is missing in the training data of the target domain.

\begin{table}[t]
  \centering
  \small
  \scalebox{0.8}{%
  \begin{tabular}{ll}
  \toprule
  Word Embedding Initialization & Correlation \\
  \midrule
  Random  &   $-$0.698 \\
  \wordtovec & $-$0.730 \\
  \wordtovec + EN & $-$0.461 \\
  \wordtovec + FS & $-$0.770 \\
  \wordtovec + ES & $-$0.514 \\
  \bottomrule
  \end{tabular}
  } %
  \caption{Correlation between in-domain data abundance and improvement from cross-domain training. The gain of cross-domain training is more significant when in-domain training data is less abundant.}
  \label{table:exp_correlation}
  \vspace{-5pt}
\end{table}

\begin{figure}[t]
  \centering
  \includegraphics[width=.7\linewidth]{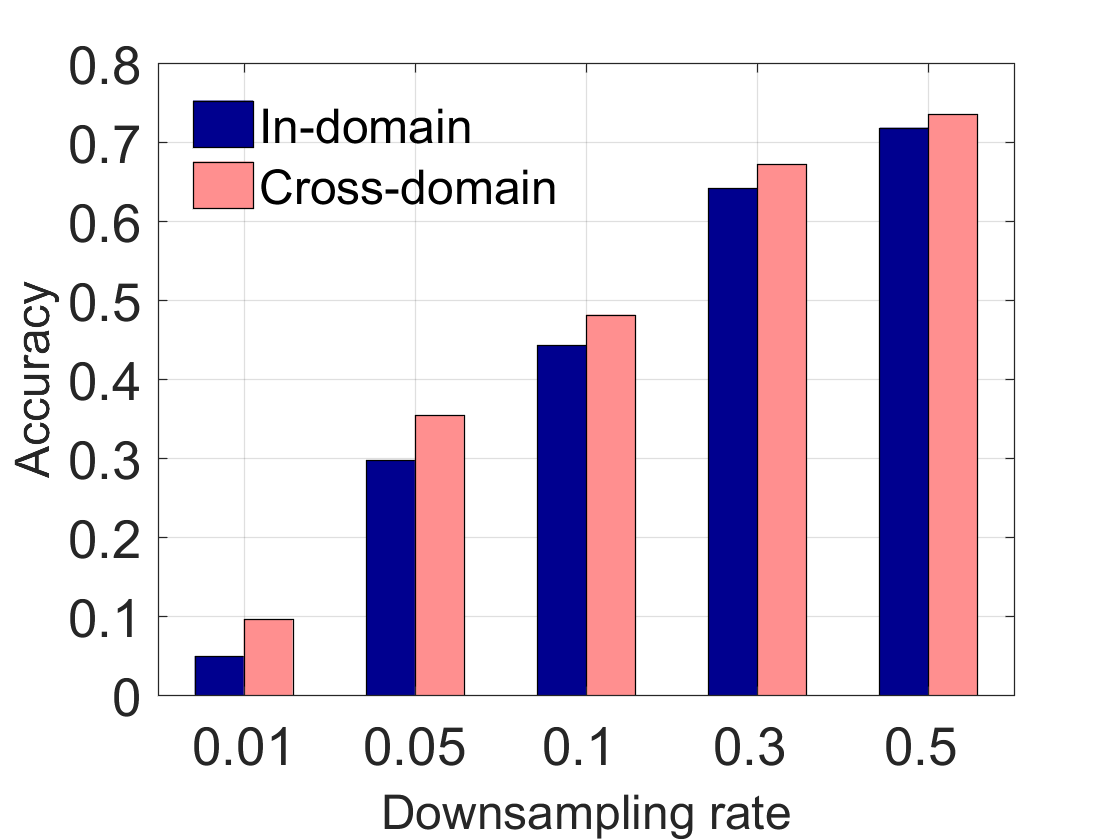}
  \vspace{-5pt}
  \caption{Results with downsampled in-domain training data. The experiment with each downsampling rate is repeated for 3 times and average results are reported. For simplicity, we only report the average accuracy over all domains. Pre-trained word embedding with per-example standardization is used in both settings.}
  \label{fig:downsample_exp}
  \vspace{-10pt}
\end{figure}

\subsubsection{Using Downsampled Training Data}
\label{sec:exp_downsample}

Compared with the vocabulary size and the number of logical forms, the in-domain training data in the \Overnight dataset is indeed abundant.
In cross-domain semantic parsing, we are more interested in the scenario where there is insufficient training data for the target domain.
To emulate this scenario, we downsample the in-domain training data of each target domain, but still use all training data from the source domain (thus $N_t \ll N_s$).
The results are shown in Figure~\ref{fig:downsample_exp}.
The gain of cross-domain training is most significant when in-domain training data is scarce.
As we collect more in-domain training data, the gain becomes smaller, which is expected.
These results reinforce those from Table~\ref{table:exp_correlation}.
It is worth noting that the effect of downsampling varies across domains.
For domains with quite abundant training data like \Social, using only 30\% of the in-domain training data, the model can achieve an accuracy almost as good as when using all the data.

\section{Discussion}
\label{sec:discussion}

Scalability, including \emph{vertical scalability}, i.e., how to scale up to handle more complex inputs and logical constructs, and \emph{horizontal scalability}, i.e., how to scale out to handle more domains, is one of the most critical challenges semantic parsing is facing today.
In this work, we took an early step towards horizontal scalability, and proposed a paraphrasing based framework for cross-domain semantic parsing.
With a sequence-to-sequence paraphrase model, we showed that cross-domain training of semantic parsing can be quite effective under a domain adaptation setting.
We also studied how to properly standardize pre-trained word embeddings in neural networks, especially for domain adaptation.

This work opens up a number of future directions.
As discussed in Section~\ref{sec:related_work}, many conventional domain adaptation and representation learning ideas can find application in cross-domain semantic parsing.
In addition to pre-trained word embeddings, other language resources like paraphrase corpora~\cite{ganitkevitch2013ppdb} can be incorporated into the paraphrase model to further facilitate domain adaptation.
In this work we require a full mapping from logical form to canonical utterance, which could be costly for large domains.
It is of practical interest to study the case where only a lexicon for mapping schema items to natural language is available.
We have restrained ourselves to the case where domains are defined using the same formal language, and we look forward to evaluating the framework on domains of different formal languages when such datasets with canonical utterances become available.

\section*{Acknowledgments}
\label{sec:acknow}

The authors would like to thank the anonymous reviewers for their thoughtful comments.
This research was sponsored in part by the Army Research Laboratory under cooperative agreements W911NF09-2-0053 and NSF IIS 1528175.
The views and conclusions contained herein are those of the authors and should not be interpreted as representing the official policies, either expressed or implied, of the Army Research Laboratory or the U.S. Government.
The U.S. Government is authorized to reproduce and distribute reprints for Government purposes notwithstanding any copyright notice herein.

\bibliography{cross_parse}
\bibliographystyle{emnlp_natbib}

\end{document}